\documentclass[twoside,11pt]{article}

\usepackage[nohyperref,preprint]{jmlr2e}

\usepackage[numbers,sort&compress]{natbib}
\usepackage[symbol]{footmisc}
\usepackage{adjustbox}
\usepackage{blindtext}
\usepackage[svgnames, table]{xcolor}
\usepackage{amsmath}        %
\usepackage{amssymb}        %
\usepackage[symbol]{footmisc}       %
\usepackage{listings}       %
\usepackage{caption}        %
\usepackage{tikz}
\usetikzlibrary{
  shadows,
  matrix,                 %
  fit,                    %
  positioning,            %
  arrows.meta,            %
  decorations.pathreplacing   %
}
\usepackage{tikz}               %
\usetikzlibrary{
  matrix,                       %
  positioning,                  %
  fit,                          %
  arrows.meta,                  %
  decorations.pathreplacing     %
}
\usepackage[frozencache,cachedir=minted-cache]{minted}
\usepackage{subcaption}
\usepackage[noend]{algpseudocode}
\usepackage{booktabs}
\usepackage{enumitem}
\usepackage[colorlinks=true,allcolors=blue]{hyperref}
\usepackage{graphicx}   %
\usepackage{svg}        %
\usepackage{tcolorbox}  %

\setminted{
  frame=single,
  framesep=2mm,
  framerule=0.4pt,
  bgcolor=blue!3,
  fontsize=\footnotesize,
  linenos=false,
  breaklines=true,
  tabsize=4,
  style=friendly,
  rulecolor=\color{gray!40}
}

\definecolor{codegreen}{rgb}{0,0.6,0}
\definecolor{codegray}{rgb}{0.5,0.5,0.5}
\definecolor{codepurple}{rgb}{0.58,0,0.82}
\definecolor{backcolour}{rgb}{0.95,0.95,0.92}

\lstdefinestyle{codeblock}{
  backgroundcolor=\color{backcolour}, commentstyle=\color{codegreen},
  keywordstyle=\color{magenta},
  numberstyle=\tiny\color{codegray},
  stringstyle=\color{codepurple},
  basicstyle=\ttfamily\footnotesize,
  breakatwhitespace=false,
  breaklines=true,
  captionpos=b,
  keepspaces=true,
  numbers=left,
  numbersep=5pt,
  showspaces=false,
  showstringspaces=false,
  showtabs=false,
  tabsize=2,
}

\usepackage{lastpage}
\firstpageno{1}

\begin{document}

\title{%
  Viser: Imperative, Web-based 3D Visualization in Python
}

\author{%
  \name Brent Yi$^{1}$ \email brentyi@berkeley.edu\\
  \vspace{-0.5cm}
  \AND
  \name Chung Min Kim$^{1}$ \email chungmin99@berkeley.edu\\
  \vspace{-0.5cm}
  \AND
  \name Justin Kerr$^{1}$ \email justin\_kerr@berkeley.edu\\
  \vspace{-0.5cm}
  \AND
  \name Gina Wu$^{1}$ \email zwu@berkeley.edu\\
  \vspace{-0.5cm}
  \AND
  \name Rebecca Feng$^{1}$ \email beckyfeng08@berkeley.edu\\
  \vspace{-0.5cm}
  \AND
  \name Anthony Zhang$^{1}$ \email anthony\_zhang1234@berkeley.edu\\
  \vspace{-0.5cm}
  \AND
  \name Jonas Kulhanek$^{2,3}$ \email jonas.kulhanek@cvut.cz\\
  \vspace{-0.5cm}
  \AND
  \name Hongsuk Choi$^{1}$ \email hongsuk@berkeley.edu\\
  \vspace{-0.5cm}
  \AND
  \name Yi Ma$^{1,4}$ \email yima@eecs.berkeley.edu\\
  \vspace{-0.5cm}
  \AND
  \name Matthew Tancik$^{5}$ \email matt@lumalabs.ai\\
  \vspace{-0.5cm}
  \AND
  \name Angjoo Kanazawa$^{1}$ \email kanazawa@eecs.berkeley.edu\\
  \vspace{-0.5cm}
  \AND
  {
  \addr $^1$UC Berkeley
  \ \ $^2$CTU in Prague
  \ \ $^3$ETH Zurich
  \ \ $^4$HKU
  \ \ $^5$Luma AI
  }
}

\maketitle

\begin{abstract}
We present Viser, a 3D visualization library for computer vision and robotics.
Viser aims to bring easy and extensible 3D visualization to Python: we provide a comprehensive set of 3D scene and 2D GUI primitives, which can be used independently with minimal setup or composed to build specialized interfaces.
This technical report describes Viser's features, interface, and implementation.
Key design choices include an imperative-style API and a web-based viewer, which improve compatibility with modern programming patterns and workflows.
Viser is open-source; code and docs can be found at \url{https://viser.studio}.

\end{abstract}

\section{Introduction}

Visual feedback enables fast, confident iteration in computer vision and robotics.
By letting researchers see, navigate, and interact with 3D data, visualization tools~\citep{Zhou2018,meshcat,hunter2007matplotlib,pyrender,pangolin,foxglove_studio,makoviychuk2021isaacgym,quigley2009ros,todorov2012mujoco,isaacsim2024,tancik2023nerfstudio,coumans2020pybullet,qin2023anyteleop,RerunSDK,abid2019gradio,sibr2020,guzov2024blendify,imgui2014} accelerate diagnosis of issues---consider incorrect coordinate conventions or self-colliding robot joints---while delivering immediate insight into both algorithmic behavior and data.

Choosing a visualization tool, however, is like many life choices. It requires weighing tradeoffs.
Most existing tools fall into one of two categories, which each have distinct advantages.
First, lightweight libraries can be imported and excel for simple visualization tasks~\cite{hunter2007matplotlib,pyrender,Zhou2018}.
They enable quick visualization with minimal setup, which is ideal for rapid debugging and prototyping.
Meanwhile, domain-specific packages offer richer features for more specific settings.
These tools are less general-purpose, but enable important workflow improvements: mobile robotics benefits from interfaces for setting 2D pose estimates~\citep{quigley2009ros}, simulators benefit from contact and force visualization~\citep{todorov2012mujoco,isaacsim2024}, and radiance fields benefit from training control and rendering interfaces~\citep{mueller2022instant,KaolinWispLibrary}.
In this work, we propose a system that aims to bridge these two categories of tools.

We present Viser, a 3D visualization library designed to be as easy as possible for simple visualization tasks in computer vision and robotics, while extending naturally to more specialized needs.
This is achieved through a comprehensive set of 3D scene and 2D GUI primitives, which can be used either independently with minimal setup or composed to build more complex, task-specific interfaces.
Design choices that improve usability include an imperative-style API and a web-based viewer, which make Viser easy to integrate with modern, Python-based programming patterns and workflows.

\begin{figure}[t!]
\centering
\includegraphics[width=\textwidth]{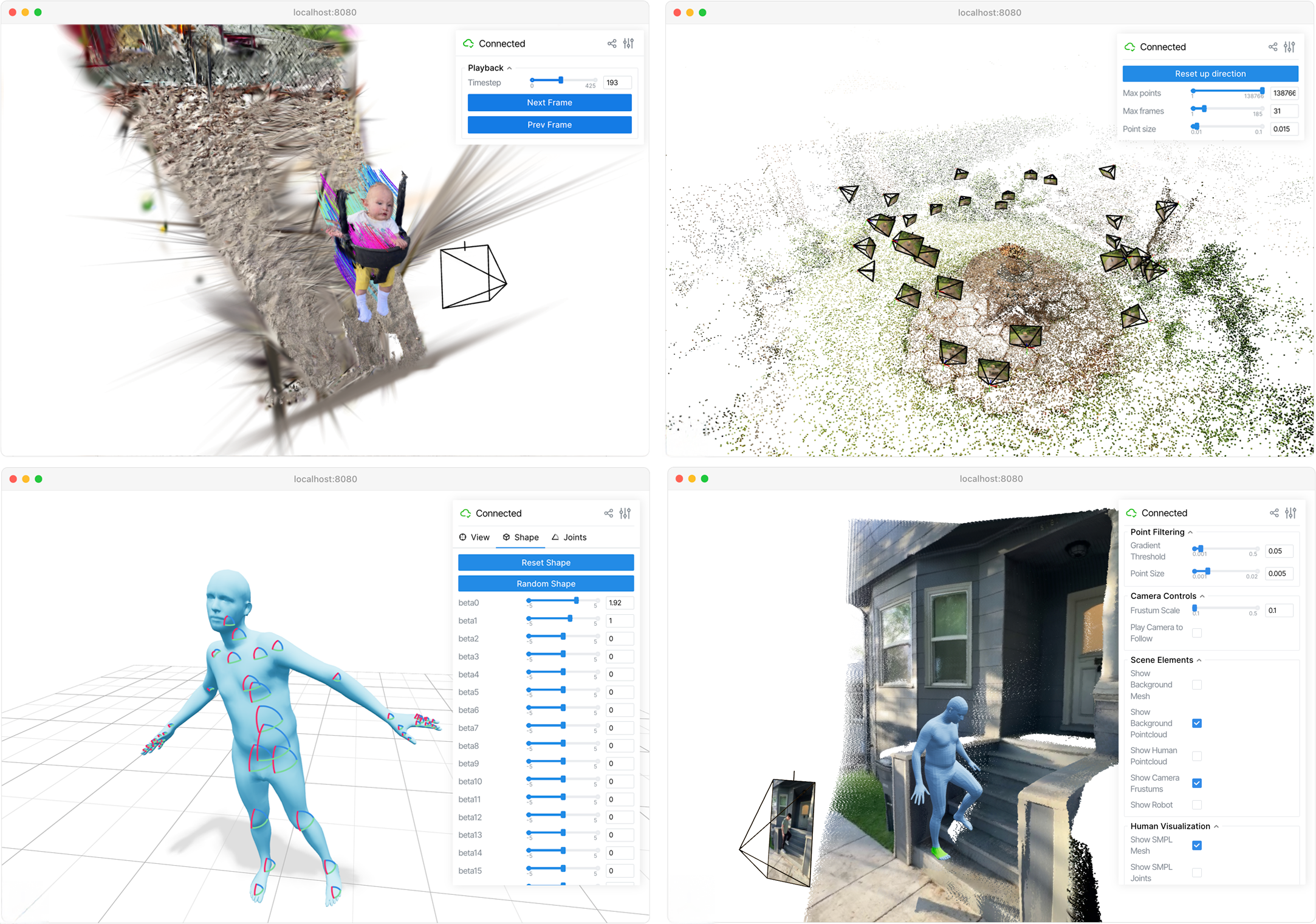}
\caption{
    \textbf{Visualization for computer vision.}
    Viser provides a web-based viewer, scene primitives, and GUI primitives for visualization.
    These can be composed for visualization in a broad set of applications.
    \textit{From top-left:} (i)~Monocular 4D reconstruction from Shape of Motion~\citep{wang2024shapeofmotion}, showing dynamic scene render and point tracks. (ii)~Visualization of mip-NeRF 360 dataset~\cite{barron2022mipnerf360} from COLMAP~\cite{schonberger2016structure}, with camera poses and point clouds. (iii)~Interactive SMPL~\citep{loper2015smpl} human model viewer with pose and shape parameter controls. (iv)~Dynamic human motion, contact, and scene visualization from VideoMimic~\citep{allshire2025videomimic}. %
    \vspace{-1em}
}
\label{fig:vision_applications}
\end{figure}

\begin{figure}[t!]
\centering
\includegraphics[width=\textwidth]{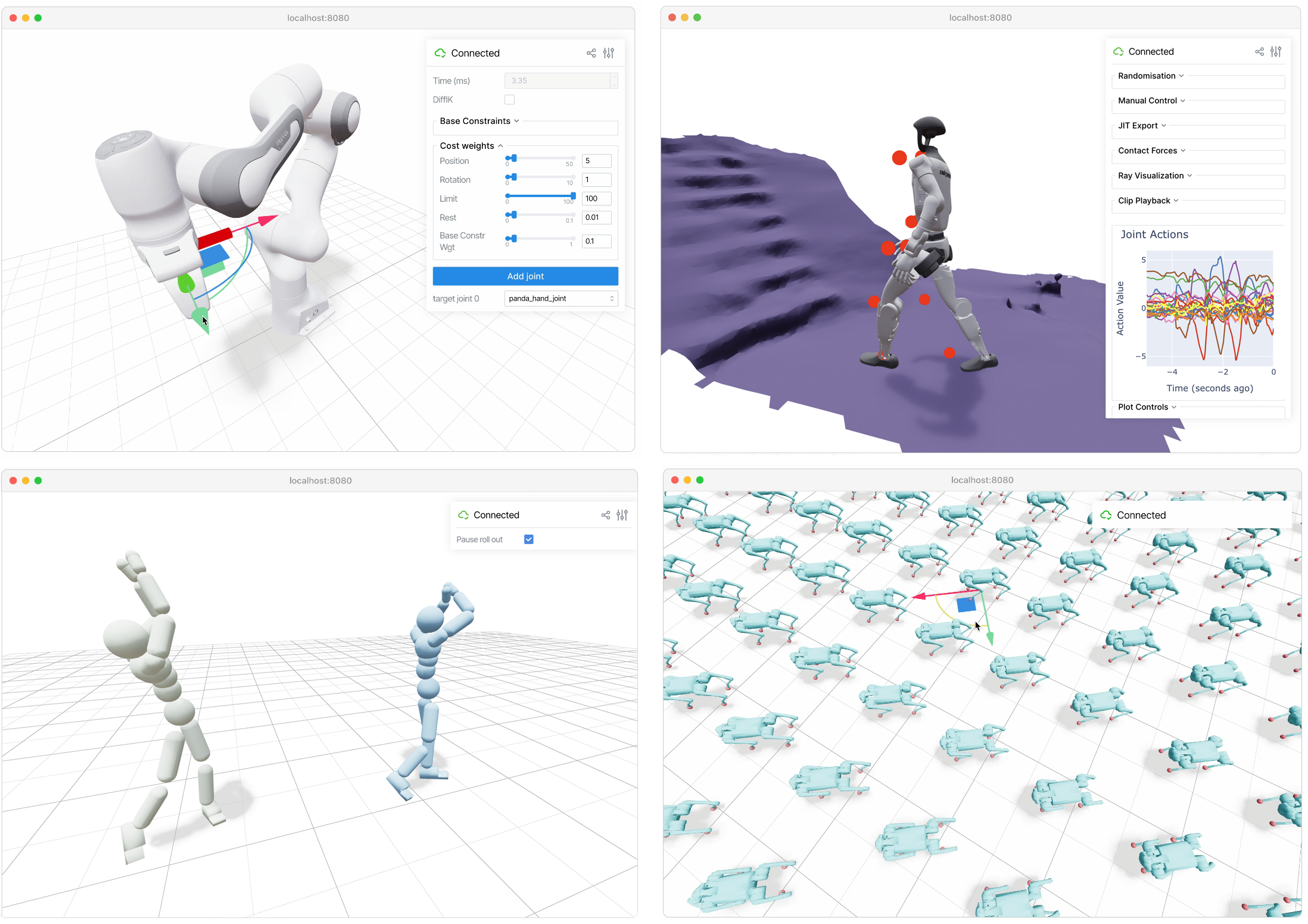}
\caption{
    \textbf{Visualization for robotics.}
    Viser's scene and GUI primitives are useful for common robotics problems.
    \textit{From top-left:} (i) Interactive inverse kinematics with 6D pose input in PyRoki~\citep{kim2025pyroki}. (ii) Policy rollout visualization for reinforcement learning in VideoMimic~\citep{allshire2025videomimic}. (iii) Humanoid control visualization from policies trained in IsaacGym~\citep{makoviychuk2021isaacgym}. (iv) Batched rendering for parallel simulation in MuJoCo Playground~\citep{zakka2025mujoco}.
    \vspace{-0.5em}
}
\label{fig:applications}
\end{figure}

\begin{figure}[t!]
\vspace{2em}
\centering
\includegraphics[width=\textwidth]{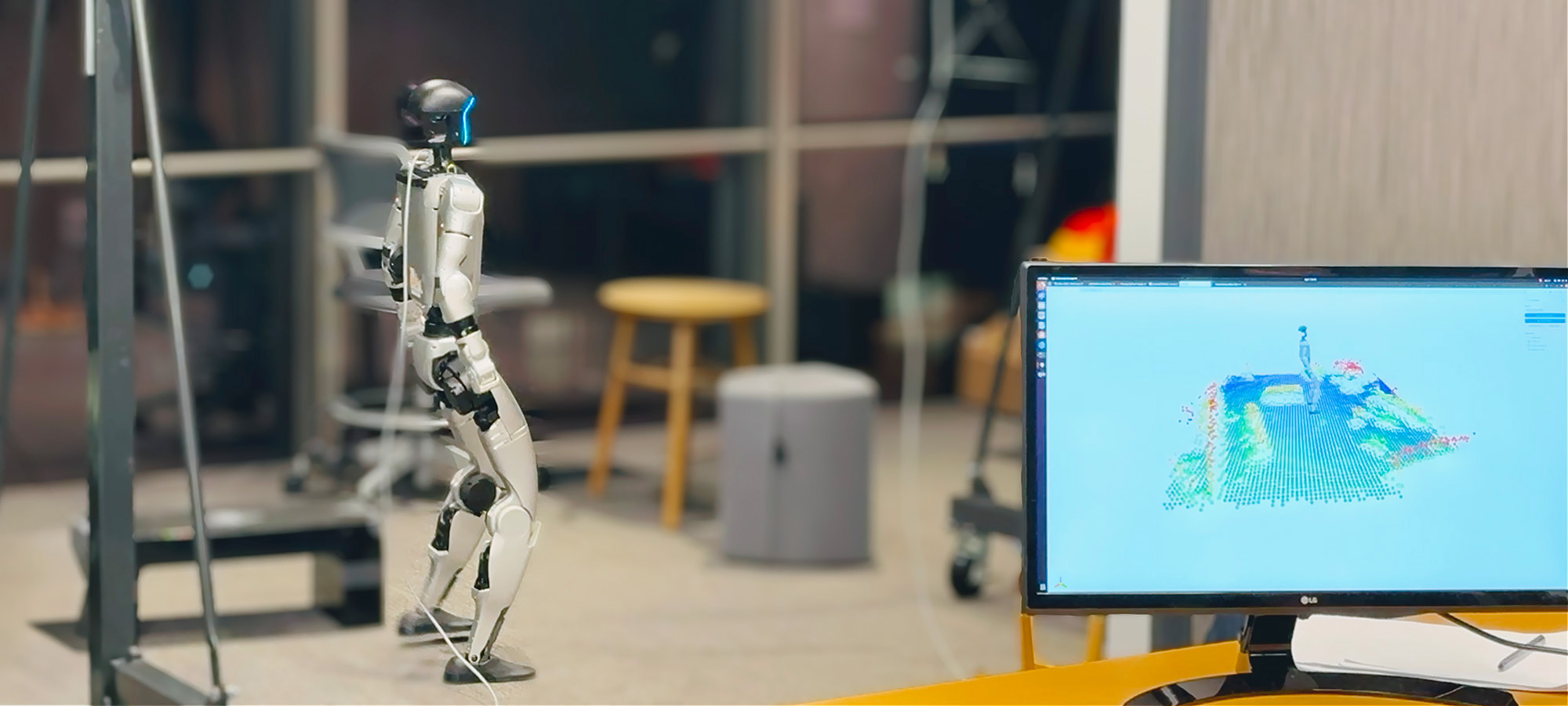}
\caption{
\textbf{Real-time visualization for robotics.}
    Viser enables live debugging of perception and control systems on physical robots. \textit{Left:} A humanoid robot executing a learned locomotion policy~\cite{allshire2025videomimic}. \textit{Right:} Visualizing real-time state estimation and mapping in Viser. Viser's web-based architecture simplifies remote monitoring and debugging. %
}
\label{fig:robot_streaming}
\end{figure}

Viser is designed to be useful for a broad set of problems.
Published works that have used Viser---for example, in released code---span neural rendering~\citep{yi2023canonical,jin2024led3dgs,kulhanek2024nerfbaselines,liu2024viewextra,ren2024scube,sinha2024spectralviewer,weber2024nerfiller,weber2024toon3d,xu2024splatfactow,ye2024gsplat,wang2024gaussianeditor,wu2025bgtriangle,neurad,tancik2023nerfstudio}, generative models~\citep{hu2024depthcrafter,liu2024viewextra,lu2024infinicube,jiang2025geo4d,weber2025fillerbuster,zhou2025seva}, scene understanding~\citep{berriel2024featsplat,gu2024egolifter,jin2024led3dgs,kim2024garfield,liu2024viewextra,adebola2025growsplat,muller2025reconstructing}, reconstruction and tracking~\citep{wang2024shapeofmotion,yang2024agenttosim,yang2024storm,yi2025egoallo,feng2025st4rtrack,jiang2025geo4d,wang2025cut3r,wu2025pod,zhang2025monst3r,yao2025uni4d,wang2025prompthmr,Yang_2025_CVPR,Wang_2025_CVPR,maggio2025vggtslamdensergbslam,he2025estimating,vasstein2025pygemini,zhang2025motion},
and robotics~\citep{qi2023rotate,rashid2023lerftogo,kerr2024rsrd,yu2024legs,allshire2025videomimic,kim2025pyroki,qi2025inhand,yu2025pogs,yu2025r2r2r,wei2024dro,bruedigam2024jacta,memmesheimer2024robocup}.
We show computer vision examples in Figure~\ref{fig:vision_applications}, as well as offline and real-time robotics examples in Figures~\ref{fig:applications} and \ref{fig:robot_streaming}.
In this paper, we begin by detailing the core features that enable these use cases (Section~\ref{sec:features}).
We then discuss Viser's underlying design: its API (Section~\ref{sec:imperative_api}) and layered system architecture~(Section~\ref{sec:implementation}).
We conclude with limitations (Section~\ref{sec:limitations}) and conclusions (Section~\ref{sec:conclusion}).

\section{Features}
\label{sec:features}

Viser proposes a unified system for 3D visualization, which is built in three key parts: a web-based viewer, a library of scene primitives, and GUI primitives.
We begin by describing these features, which are designed to simplify visualization across computer vision (Figure~\ref{fig:vision_applications}) and robotics (Figure~\ref{fig:applications}) tasks.

\subsection{Web-based Viewer}
Viser automatically hosts a local visualization server when used, which provides a viewer that can be accessed from any modern web browser.
This web-based approach has several advantages.
\textit{(1)~Low setup effort:}
a web-based viewer makes Viser simple to install and run across platforms, including on headless servers and mobile clients.
\textit{(2)~Sharing:}
Viser lets researchers embed offline visualizations into static webpages or share active sessions using simple URLs.
Embedding examples can be found on webpages for \citep{yi2025egoallo,wang2025cut3r,wang2024shapeofmotion,zhang2025monst3r,kerr2024rsrd}.
\textit{(3)~Development:}
web infrastructure accelerates development velocity---Viser benefits from mature libraries like React~\citep{react2013} and \texttt{three.js} for UI development and 3D rendering.

\begin{figure}[t!]
\centering
\includegraphics[width=\textwidth]{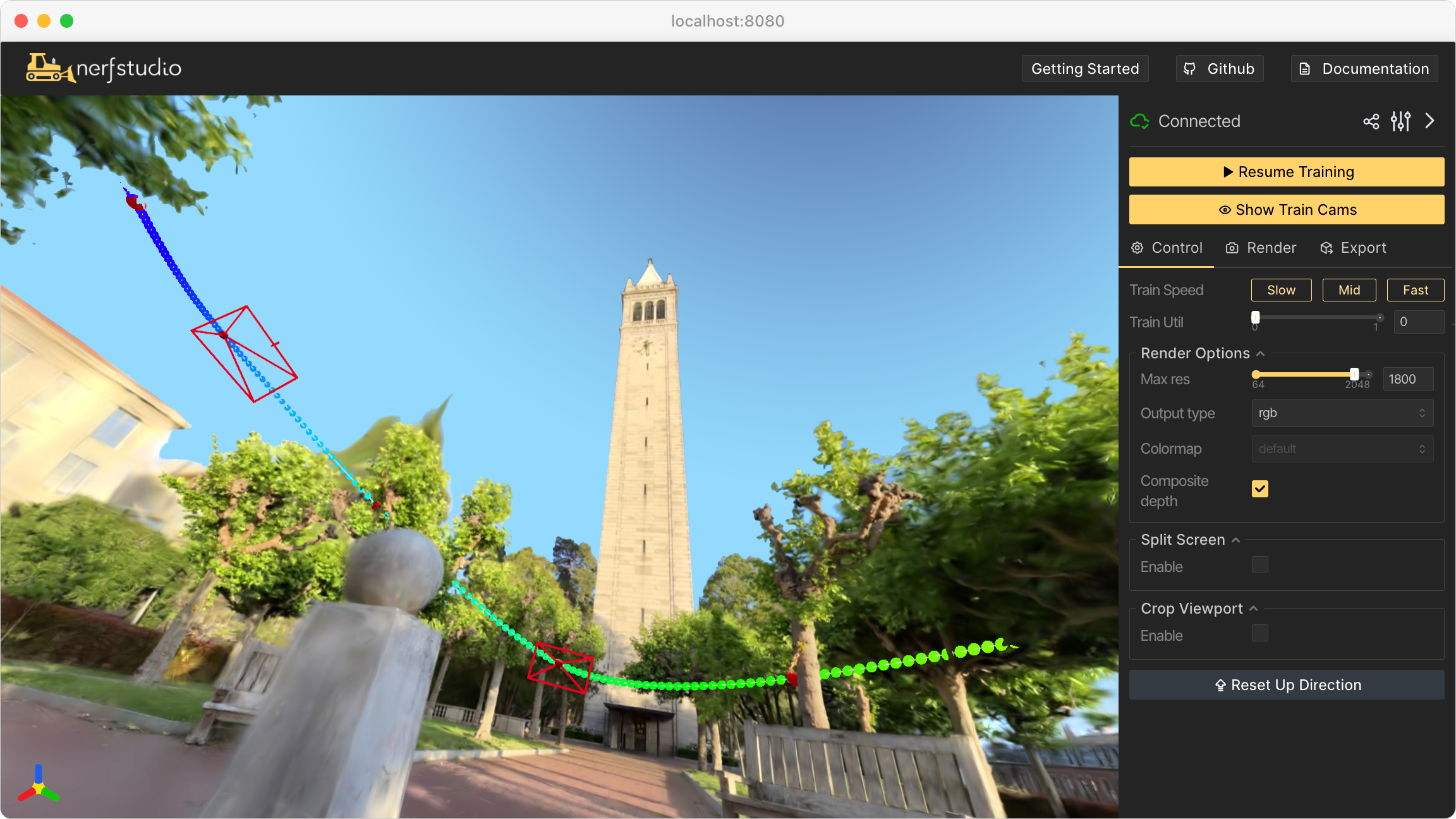}
\caption{
    \textbf{Nerfstudio~\citep{tancik2023nerfstudio}.}
    An example of a domain-specific tool built with Viser's scene and GUI primitives. The viewer supports real-time rendering of neural radiance fields~\cite{mildenhall2020nerf} and 3D Gaussian splats~\cite{kerbl2023gaussian}, training visualization, and camera path creation. %
}
\label{fig:nerfstudio}
\end{figure}

\subsection{Scene Primitives}
\label{sec:capabilities_3d}

We implement a comprehensive range of 3D visualization features, which aim to be generally useful across applications:

\textbf{Visualizing diverse 3D data.}
Users can display various data types---point clouds, meshes, images, Gaussian splats---along with geometric primitives like coordinate frames, frustums, grids, splines, and line segments, all with simple function calls.
Viser can directly load GLB and glTF formats, which enables integration with existing 3D assets.
Batched rendering and level-of-detail optimizations maintain performance for large scenes.

\textbf{Organizing complex scenes.} Viser uses a hierarchical scene graph that simplifies coordinate transformations and kinematic relationships. This generalizes nested coordinate systems like robots and camera rigs.

\textbf{Creating high-quality visuals.}
Viser's \texttt{three.js}-based rendering pipeline supports physically-based materials, environment maps, comprehensive lighting (ambient, directional, point, area, spot), and shadow mapping. These features enable high-quality visuals without exporting to specialized rendering software.

\textbf{Handling real-time data streams.} 
Viser automatically synchronizes state changes between Python and web clients. Updates are optimized to enable smooth visualization of dynamic data from neural network training, physics simulations, and robot sensors.

\textbf{Building interactive applications.} To move beyond passive visualization, users can make objects clickable to trigger events, add transform gizmos to specify poses, and subscribe to general scene pointer events. A camera API enables programmatic access to and control over viewpoint information. %

\subsection{GUI Primitives}
\label{sec:capabilities_2d}

Viser also provides 2D GUI features.
The 2D GUI enables domain-specific controls and custom information displays:

\textbf{Capturing user input.} Users can create standard interface elements---buttons, sliders, checkboxes, text inputs, dropdowns---with single function calls like \texttt{gui.add\_button()}. Specialized controls like color pickers, vector inputs, upload buttons, and progress bars can be used for more advanced applications.

\textbf{Displaying rich information.} Beyond inputs, users can render text, markdown, and HTML content inline with visualizations. 2D images can be both displayed and streamed, while Plotly~\citep{plotly2015} and uPlot~\cite{uPlot_1.6.32} integration enables 2D plotting. A notification system provides status updates and user feedback.

\textbf{Organizing complex interfaces.} As applications grow, folders and tab groups help structure complicated control panels, while modal dialogs handle focused interactions.
GUI containers can also be placed directly in 3D scenes, creating spatially integrated inputs. This enables specialized use cases like per-object control panels and annotation tooling.

\begin{figure}[htbp]
    \vspace{-0.02em}
    \centering
    \begin{subfigure}[b]{\textwidth}
        \centering
        \includegraphics[width=\textwidth]{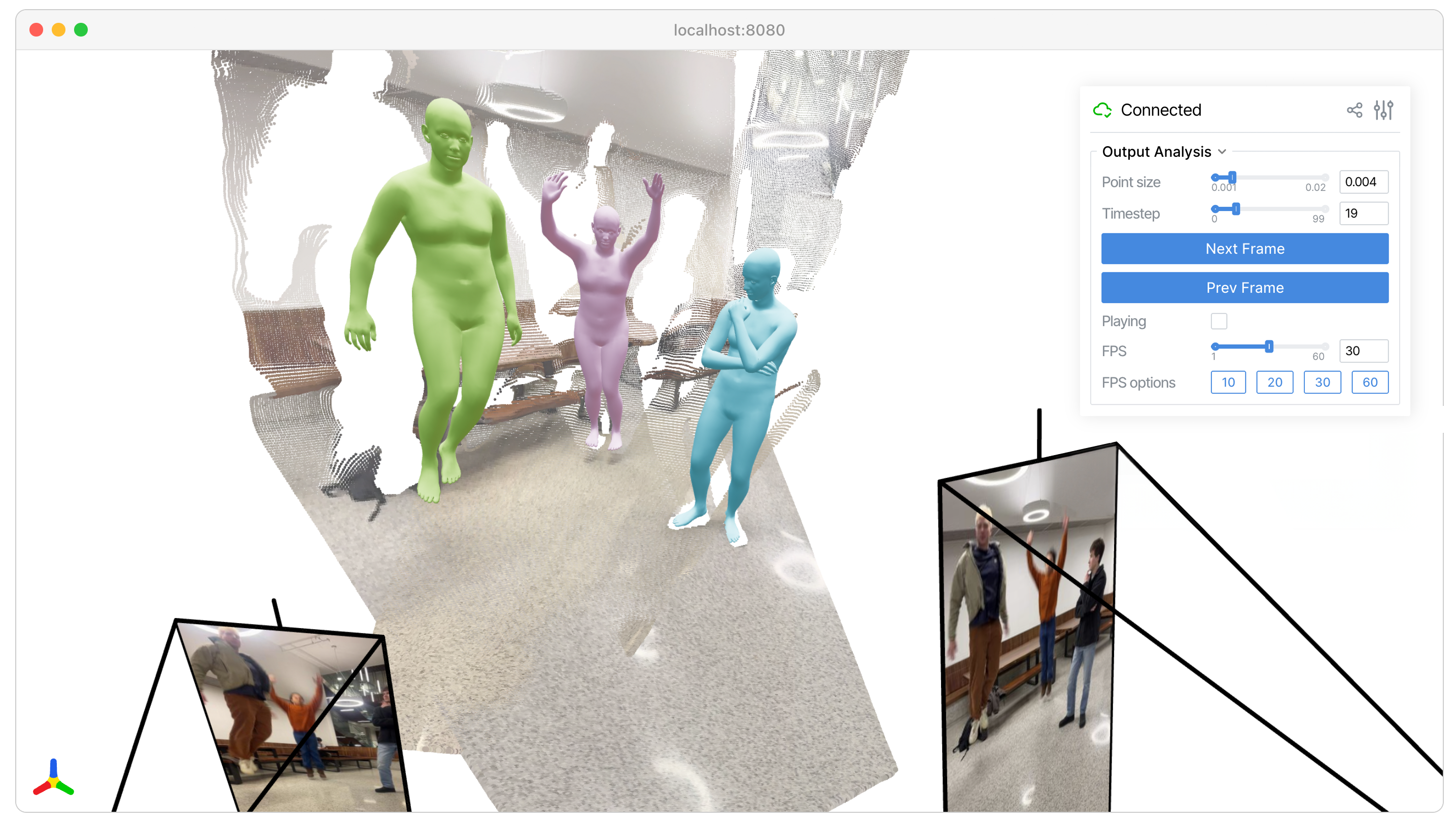}
        \vspace{-0.5em}
        \caption{\textit{Web-based client, with the scene and user interface specified in Python}}
        \label{fig:overview_client}
    \end{subfigure}
    \begin{subfigure}[t]{0.48\textwidth}
        \centering
        \begin{minted}{python}
scene.add_point_cloud(
  "/points",
  points=np.array(...),
  colors=np.array(...),
)

scene.add_camera_frustum(
  "/camera",
  fov=np.pi / 2.0,
  aspect=h / w,
  wxyz=(qw, wx, qy, qz),
  position=(x, y, z),
)

# Add other camera frustums, meshes
        \end{minted}
        \caption{\textit{Python API for scene control}}
        \label{fig:overview_scene_api}
    \end{subfigure}
    \hfill
    \begin{subfigure}[t]{0.48\textwidth}
        \centering
        \begin{minted}{python}
next_btn = gui.add_button("Next Frame")
prev_btn = gui.add_button("Prev Frame")
playing_cb = gui.add_checkbox("Playing")

@next_btn.on_click
def _(_) -> None:
    ...
@prev_btn.on_click
def _(_) -> None:
    ...
@playing_cb.on_update
def _(_) -> None:
    ...

# Add other UI elements
        \end{minted}
        \caption{\textit{Python API for GUI configuration}}
        \label{fig:overview_gui_api}
    \end{subfigure}
    \caption{
        \textbf{Populating Viser.}
        Viser is used to display inputs and outputs from a multiview reconstruction pipeline~\cite{muller2025reconstructing}.
        We show (a) the viewer, (b) example code for adding 3D scene elements, and (c) example code for populating the graphical interface.
    }
    \label{fig:viser_overview}
\end{figure}

\vspace{3em} %
\section{API}
\label{sec:imperative_api}

Viser's API design is centered on ease-of-use and flexibility: in practice, this means alignment with standard Python programming patterns.
We design Viser's scene and GUI features to integrate seamlessly with arbitrary Python programs, notebooks~\citep{kluyver2016jupyter}, REPLs~\citep{van1995python,perez2007ipython}, and step-through debuggers.
To achieve this, we propose an \textit{imperative-style} API characterized by explicit side effects in user-controlled program flow. %

\begin{figure}[t!]
\centering
\begin{subfigure}[b]{0.48\textwidth}
    \begin{minted}{python}
# Create box.
box = scene.add_box("/box")

# Update box property.
box.color = (255, 0, 0)

# Attach click callback.
@box.on_click
def _(event): print("Box clicked")

# Remove box.
box.remove()
    \end{minted}
    \caption{3D Scene API}
\end{subfigure}
\hfill
\begin{subfigure}[b]{0.48\textwidth}
    \begin{minted}{python}
# Create button.
button = gui.add_button("Click")

# Update button property.
button.color = (255, 0, 0)

# Attach click callback.
@button.on_click
def _(event): print("Button clicked")

# Remove button.
button.remove()
    \end{minted}
    \caption{2D GUI API}
\end{subfigure}
\caption{
    \textbf{Imperative, side effect-driven component lifecycle management.}
    The same programming patterns apply to both 3D scene and 2D GUI primitives: creating objects with single function calls, updating properties through handle attributes, registering event callbacks with Python decorators, and explicit removal when no longer needed.
}
\label{fig:lifecycle_management}
\end{figure}

\textbf{Primitive lifecycle.}
Scene and GUI elements in Viser are created through single function calls, which return handles for subsequent interaction (Figure~\ref{fig:viser_overview},~\ref{fig:lifecycle_management}).
These handles encapsulate both lifecycle control and automatic state synchronization: users manage when primitives are removed, and can use assignment-style syntax to update properties.
Property assignments immediately update visualizations, without manual synchronization or render calls.
Similarly, user interactions in the browser---clicks, slider movements, text input---automatically flow back to Python, where they are made available for both polling-style reads and interrupt-style callbacks. %
This design combines explicit control over object lifecycles with implicit handling of the logic required for real-time, bidirectional synchronization between Python and web clients.

\begin{figure}[t!]
\centering

\begin{subfigure}[b]{0.48\textwidth}
    \begin{minted}{python}
# Create elements.
slider = gui.add_slider("Value", 0, 100)
out = gui.add_text("0")

# Direct state update on event.
@slider.on_update
def _(_):
    out.value = str(slider.value * 2)

# Continue with other Python code.
while True:
    time.sleep(1.0)
    \end{minted}
    \vspace{-2em}
    \caption{Imperative input/output relationships.}
\end{subfigure}
\hfill
\begin{subfigure}[b]{0.48\textwidth}
    \begin{minted}[bgcolor=orange!5]{python}
# Define UI as input-output transform.
def double_value(x):
    return str(x * 2)

demo = gr.Interface(
    fn=double_value,
    inputs=gr.Slider(0, 100),
    outputs=gr.Textbox(),
)

# Framework takes control.
demo.launch()
    \end{minted}
    \vspace{-2em}
    \caption{Declarative input/output relationships.}
\end{subfigure}

\begin{subfigure}[b]{0.5\textwidth}
    \begin{minted}{python}
# Explicit state management.
counter = 0
label = gui.add_text(f"Count: {counter}")
button = gui.add_button("Increment")

# Update state on click.
@button.on_click
def _(_):
    global counter
    counter += 1
    label.value = f"Count: {counter}"

# Continue with other Python code.
while True:
    time.sleep(1.0)
    \end{minted}
    \vspace{-2em}
    \caption{Imperative state management.}
\end{subfigure}
\hfill
\begin{subfigure}[b]{0.48\textwidth}
    \begin{minted}[bgcolor=orange!5]{python}
# Framework manages state.
def increment(count):
    return (
        count + 1,
        f"Count: {count + 1}",
    )

demo = gr.Interface(
    fn=increment,
    inputs=[gr.State(0), gr.Button()],
    outputs=[gr.State(), gr.Textbox()],
)

# Framework takes control.
demo.launch()
    \end{minted}
    \vspace{-2em}
    \caption{Declarative state management.}
\end{subfigure}

\caption{
    \textbf{Comparing \protect\colorbox{blue!7}{imperative} and \protect\colorbox{orange!7}{declarative} abstractions.}
    Viser's imperative-style API provides low-level control through side effects, which are replaced with higher-level abstractions in declarative APIs like Gradio~\cite{abid2019gradio}, the dominant library for building user interfaces for machine learning applications.
    Viser's imperative-style API prioritizes user control of program flow, which simplifies integration into complex programs.
    In this Viser example, state is shared between all clients; per-client state management is also possible.
}
\label{fig:imperative_vs_declarative}
\end{figure}

\textbf{Why imperative?}
Viser's imperative-style design deviates from the declarative patterns that are more common for layout and visualization in Python~\citep{abid2019gradio,streamlit2020,schindler2025nicegui,plotly2015,satyanarayan2017vega,vanderplas2018altair}.
In declarative programming, users pass objectives like visual properties, data relationships, and scene layouts to a runtime, which is then responsible for the details of program flow.
This simplifies many use cases: Gradio~\cite{abid2019gradio}, for example, excels at managing independent state for multiple parallel clients.
This comes at a cost: handing off control over program flow can limit integration flexibility. %
In contrast, Viser's imperative programming model is designed to be easier to integrate into interactive Python programming patterns.
We provide a concrete comparison in Figure~\ref{fig:imperative_vs_declarative}.

\section{System Architecture}
\label{sec:implementation}

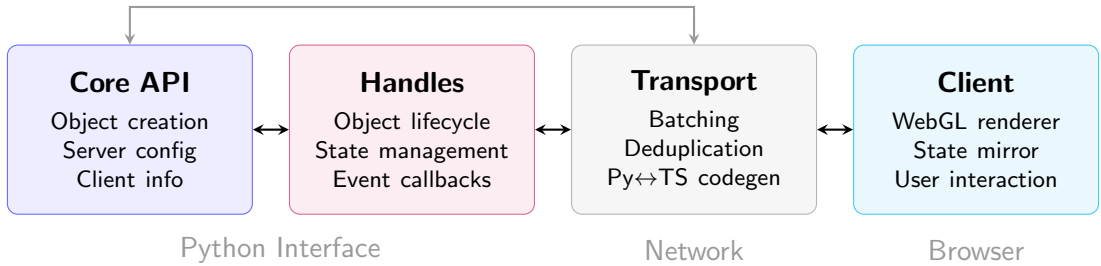
\begin{figure}[t]
\centering
\begin{tikzpicture}[
    layer/.style={
        rectangle,
        rounded corners=4pt,
        minimum width=3.2cm,
        minimum height=2.25cm,
        text width=3cm,
        align=center,
        font=\sffamily,
    },
    arrow/.style={
        <->,
        thick,
        >=stealth,
    }
]

\node[layer, fill=blue!7, draw=blue!60] (api) at (-5.5,0) {
    \textbf{Core API} \\[0.1cm]
    {
        \footnotesize
        Object creation\\
        Server config\\
        Client info\\
    }
};

\node[layer, fill=purple!7, draw=purple!60] (handle) at (-1.75,0) {
    \textbf{Handles} \\[0.1cm]
    {
        \footnotesize
        Object lifecycle\\
        State management\\
        Event callbacks\\
    }
};

\node[layer, fill=gray!7, draw=gray!60] (transport) at (2,0) {
    \textbf{Transport} \\[0.1cm]
    {
        \footnotesize
        Batching\\
        Deduplication\\
        Py$\leftrightarrow$TS codegen\\
    }
};

\node[layer, fill=cyan!7, draw=cyan!60] (client) at (5.75,0) {
    \textbf{Client} \\[0.1cm]
    {
        \footnotesize
        WebGL renderer\\
        State mirror\\
        User interaction\\
    }
};

\draw[arrow] (api.east) -- (handle.west);
\draw[arrow] (handle.east) -- (transport.west);
\draw[arrow] (transport.east) -- (client.west);
\draw[arrow,black!40] (api.north) -- ++(0,0.5) -- ++(7.5,0) -- (transport.north);

\node[font=\sffamily, text=gray!80] at (-3.5, -1.6) {Python Interface};
\node[font=\sffamily, text=gray!80] at (2.0, -1.6) {Network};
\node[font=\sffamily, text=gray!80] at (5.75, -1.6) {Browser};

\end{tikzpicture}
\caption{
    \textbf{Implementation overview.}
    Viser is implemented in four layers.
    Users interact programmatically with a Python interface (Core API, Handles), which communicate (Transport) with a web-based frontend (Client).
}
\label{fig:layers}
\end{figure}

To support the imperative-style API described in Section~\ref{sec:imperative_api}, Viser employs an architecture that separates user-facing interfaces from the underlying web-based implementation (Figure~\ref{fig:layers}).
This design shields users from networking and state synchronization complexities.
It is implemented in four layers.

\textbf{Core API.}
Viser's core API provides high-level methods like \texttt{scene.add\_mesh()} and \texttt{gui.add\_button()} for creating objects, as well as methods for configuring servers and accessing client information.
These methods can be called either globally to share objects between all clients or on a per-client basis for more targeted updates.

\textbf{Handles.} Each primitive creation method call returns a handle that maintains the created element's lifecycle and state.
Handles provide property setters and getters, where operations like \texttt{box.color = (255, 0, 0)} immediately update the visualization.
They also offer event callback registration for building interactive applications.

\textbf{Transport.}
Commands from Viser's Python-based API are sent to clients through Viser's transport layer, which manages communication between Python and web clients via WebSocket connections.
The transport layer automatically buffers, batches, and deduplicates state updates in a thread before serializing them via msgpack~\citep{furuhashi2008msgpack}.
Python message definitions are used to generate TypeScript declarations for static verification, ensuring type safety across the Python-JavaScript API boundary.
Deduplicated messages are efficiently persisted to maintain state consistency for new clients.
Latency-aware buffering enables smooth framerates, even on unreliable network connections.

\textbf{Client.} Clients receive updates and render them in a web browser.
Viser clients maintain a mirror of server-side state and capture user interactions---clicks, camera motion, and GUI input changes---which are relayed back to the Python server through the same transport layer.
Performance optimizations include threading via WebWorkers, Gaussian splat sorting compiled via WebAssembly, and shadows rendered with cascaded shadow maps~\cite{Dimitrov2007CSM}.

\section{Limitations}
\label{sec:limitations}

While Viser's design and implementation provides advantages in simplicity and flexibility, achieving these benefits requires tradeoffs.
Current limitations include:
\begin{itemize}[itemsep=1pt]
    \item \textbf{WebSocket transfer.} Viser's web-based client receives all visualized assets through a WebSocket connection. This introduces overhead, which is exacerbated by the fact that program state is managed by the Python server: updates in response to user interaction require round-trip messages and cannot be exported to static webpages. Websocket-based message passing also prevents CPU-to-GPU transfer optimizations like the ones implemented in~\citet{teed2023deep}.
    \item \textbf{Stateful API.} Viser's API is inherently stateful. This is an intentional design choice, but can also result in more duplicated or error-prone state management than declarative~\citep{abid2019gradio} or immediate-mode~\citep{imgui2014} visualization APIs.
    \item \textbf{Python.} Viser is designed for Python users. We do not provide bindings for languages like C++ or Rust, which are common in performance-critical systems.
    \item \textbf{Single process.} Viser launches per-script visualization servers.
    This may require effort to adapt to systems with concurrent processes, which are common in robotics~\citep{quigley2009ros}.
    \item \textbf{Timestamps.} Viser does not assume temporal structure or provide standard ways to timestamp data.
    This can increase boilerplate for visualizing sequence data.
    \item \textbf{Serialization.} Many tools can load data from formats like rosbag~\cite{quigley2009ros}, MCAP~\cite{foxglove_studio}, and rrd~\cite{RerunSDK}.
    Serialized data is useful for logging and offline playback, which Viser has limited built-in features for.
\end{itemize}

\section{Conclusion}
\label{sec:conclusion}
We presented Viser, a Python library for 3D visualization.
Viser provides 3D scene and 2D GUI visualization primitives that aim to be easy to use independently, but are powerful when composed.
We hope that Viser will be useful for accelerating research and engineering efforts in computer vision, robotics, and related fields.

\section{Acknowledgements}

Viser is an open-source project that has been made possible by a community of contributors:
we thank
Hang Gao, 
Jonah Bedouch, 
Brian Santoso,
Abhik Ahuja,
Ethan Weber,
Sebastian Castro,
Adam Rashid,
zerolover,
Jihoon Oh,
Sylvain Poulain,
Arthur Allshire,
Simon Le Cleac'h,
Albert Li,
Karan Desai,
Alejandro Escontrela,
Zhensheng Yuan,
Adam Tonderski,
Simon Bethke,
Jonathan Zakharov,
Rohan Mathur,
Liam Schoneveld,
Cyrus Vachha,
David McAllister,
Andrea Boscolo Camiletto,
and Christian Le for contributing features and bug fixes.
We would also like to thank Sam Buchanan and Weijia Zeng for discussions during Viser's initial development~\cite{yi2023canonical},
Lea M\"uller, Qianqian Wang, Haiwen Feng, and Junyi Zhang for software feedback, and Vickie Ye for critical implementation and design discussions.

Viser's design is heavily influenced by existing software packages: notable inspirations include Pangolin~\cite{pangolin}, Dear ImGui~\cite{imgui2014}, Meshcat~\cite{meshcat}, rviz~\cite{quigley2009ros}, and Gradio~\cite{abid2019gradio}.
We thank the authors of these packages for open-sourcing their work.

\textit{Funding.}
Brent Yi, Chung Min Kim, and Justin Kerr are supported by the NSF Graduate Research Fellowship Program, Grant DGE 2146752.

\newpage

\vskip 0.2in
\bibliography{references}

\newpage

\end{document}